\documentclass{article}

\usepackage{microtype}
\usepackage{graphicx}
\usepackage{subcaption}
\usepackage{booktabs} % for professional tables
\usepackage{multirow}

\usepackage{hyperref}

\usepackage[preprint]{icml2026}

\usepackage{amsmath}
\usepackage{amssymb}
\usepackage{mathtools}
\usepackage{amsthm}

\usepackage[capitalize,noabbrev]{cleveref}

\theoremstyle{plain}

\theoremstyle{definition}

\theoremstyle{remark}

\usepackage[textsize=tiny]{todonotes}

\icmltitlerunning{SyncAnyone: Implicit Disentanglement via Progressive Self-Correction for Lip-Syncing in the wild}

\begin{document}

\twocolumn[
  \icmltitle{SyncAnyone: Implicit Disentanglement via Progressive Self-Correction \\
    for Lip-Syncing in the wild}

  % It is OKAY to include author information, even for blind submissions: the
  % style file will automatically remove it for you unless you've provided
  % the [accepted] option to the icml2026 package.

  % List of affiliations: The first argument should be a (short) identifier you
  % will use later to specify author affiliations Academic affiliations
  % should list Department, University, City, Region, Country Industry
  % affiliations should list Company, City, Region, Country

  % You can specify symbols, otherwise they are numbered in order. Ideally, you
  % should not use this facility. Affiliations will be numbered in order of
  % appearance and this is the preferred way.
  \icmlsetsymbol{equal}{*}

  \begin{icmlauthorlist}
    \icmlauthor{Xindi Zhang}{equal,comp}
    \icmlauthor{Dechao Meng}{equal,comp}
    \icmlauthor{Steven Xiao}{equal,comp}
    \icmlauthor{Qi Wang}{comp}
    \icmlauthor{Peng Zhang}{comp}
    \icmlauthor{Bang Zhang}{comp}
  \end{icmlauthorlist}
  
  % Xindi Zhang\footnotemark[1] \quad Dechao Meng\footnotemark[1] \quad Steven Xiao\footnotemark[1] \\ \quad Qi Wang \quad Peng Zhang \quad Bang Zhang \\
  %   Tongyi Lab, Alibaba Group 

  % \icmlaffiliation{yyy}{Department of XXX, University of YYY, Location, Country}
  \icmlaffiliation{comp}{Tongyi Lab, Alibaba Group}
  % \icmlaffiliation{sch}{School of ZZZ, Institute of WWW, Location, Country}

  \icmlcorrespondingauthor{Xindi Zhang}{zhangxindi.zxd@alibaba-inc.com}
  \icmlcorrespondingauthor{Dechao Meng}{mengdechao.mdc@alibaba-inc.com}

  % You may provide any keywords that you find helpful for describing your
  % paper; these are used to populate the "keywords" metadata in the PDF but
  % will not be shown in the document
  \icmlkeywords{Machine Learning, ICML}

  \vskip 0.3in
]

% this must go after the closing bracket ] following \twocolumn[ ...

% This command actually creates the footnote in the first column listing the
% affiliations and the copyright notice. The command takes one argument, which
% is text to display at the start of the footnote. The \icmlEqualContribution
% command is standard text for equal contribution. Remove it (just {}) if you
% do not need this facility.

% Use ONE of the following lines. DO NOT remove the command.
% If you have no special notice, KEEP empty braces:
\printAffiliationsAndNotice{}  % no special notice (required even if empty)
% Or, if applicable, use the standard equal contribution text:
% \printAffiliationsAndNotice{\icmlEqualContribution}

\begin{abstract}

High-quality AI-powered video dubbing demands precise audio-lip synchronization, high-fidelity visual generation, and faithful preservation of identity and background. Most existing methods rely on a mask-based training strategy, where the mouth region is masked in talking-head videos, and the model learns to synthesize lip movements from corrupted inputs and target audios. While this facilitates lip-sync accuracy, it disrupts spatiotemporal context, impairing performance on dynamic facial motions and causing instability in facial structure and background consistency. To overcome this limitation, we propose SyncAnyone, a novel two-stage learning framework that achieves accurate motion modeling and high visual fidelity simultaneously. 
In Stage 1, we train a diffusion-based video transformer for masked mouth inpainting, leveraging its strong spatiotemporal modeling to generate accurate, audio-driven lip movements. However, due to input corruption, minor artifacts may arise in the surrounding facial regions and the background. In Stage 2, we develop a mask-free tuning pipeline to address mask-induced artifacts. Specifically, on the basis of the Stage 1 model, we develop a data generation pipeline that creates pseudo-paired training samples by synthesizing lip-synced videos from the source video and random sampled audio. 
We further tune the stage 2 model on this synthetic data, achieving precise lip editing and better background consistency.
Extensive experiments show that our method achieves state-of-the-art results in visual quality, temporal coherence, and identity preservation under in-the wild lip-syncing scenarios. Project page: \url{https://humanaigc.github.io/sync_anyone_demo_page/}.

\end{abstract}    
\section{Introduction}
\begin{figure}
    \centering
    \includegraphics[width=\linewidth]{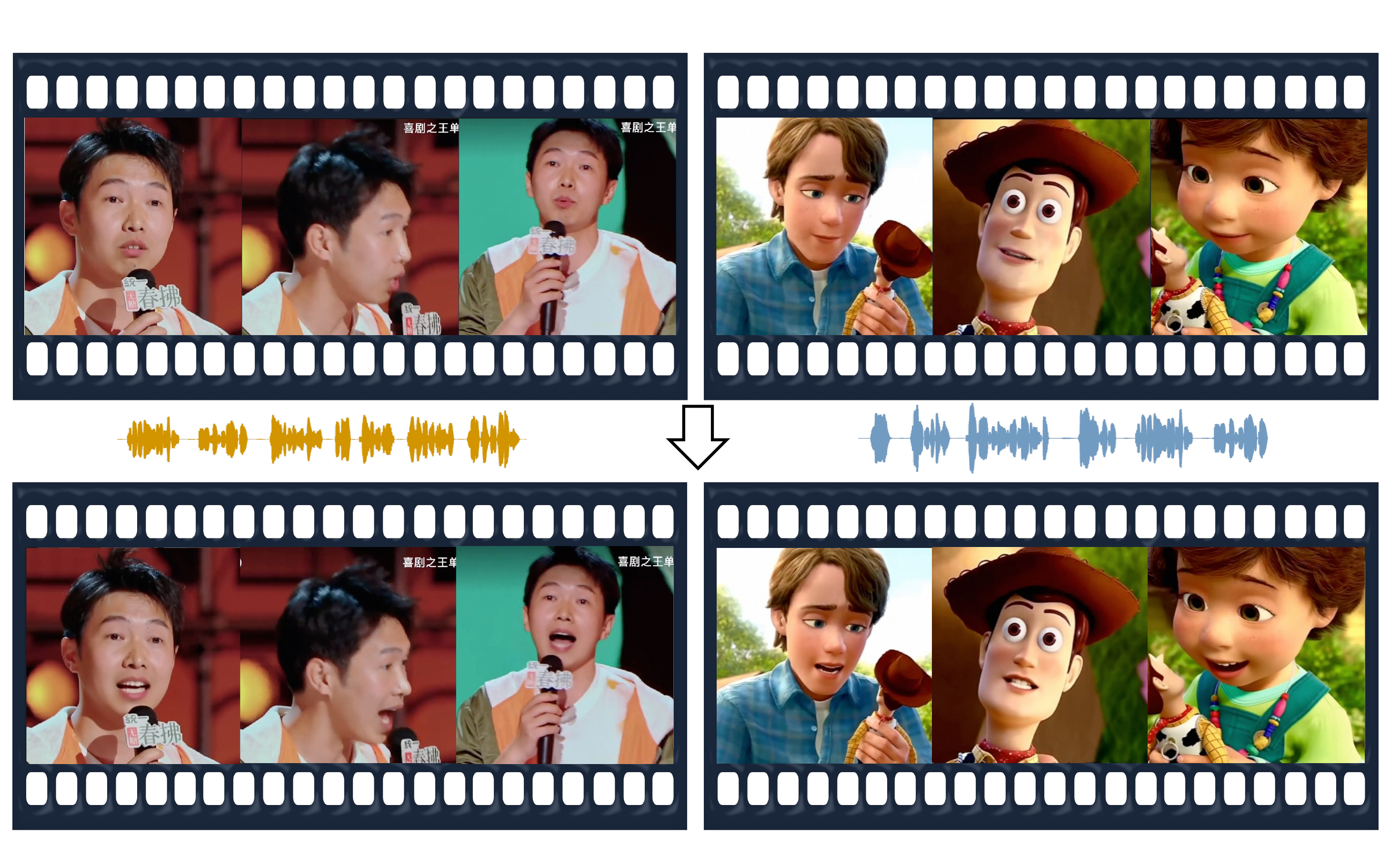}
    \caption{Given any video and audio input, SyncAnyone can modify the mouth movements of characters in the video to synchronize with the audio. Additionally, SyncAnyone is capable of handling challenging scenarios such as large poses, background changes, occlusions, scene cuts, and diverse styles.
}
    \label{fig:intro}
\end{figure}
\label{sec:introduction}

Audio-driven lip synchronization, or lip-syncing, is the task of algorithmically modifying a speaker's mouth in a video to achieve precise synchronization with a target audio signal. A key requirement of this task is to preserve the subject's identity, non-verbal facial expressions, and the integrity of the background scene. This technology has significant application potential in domains such as post-production video dubbing, virtual avatar animation, online education, and cross-lingual content localization. An ideal lip-syncing system is expected to not only synthesize photorealistic and temporally coherent mouth movements but also exhibit robustness against common in-the-wild challenges (Fig.~\ref{fig:intro}), including extreme head poses, partial occlusions, scene cuts, and complex background dynamics.

Recently, diffusion transformers (DiTs) have emerged as a promising approach to lip-syncing. By leveraging strong spatiotemporal coherence priors from large-scale video data, mainstream methods adopt a mask inpainting paradigm: they sample video sequences with the mouth region masked and train models to reconstruct the lips conditioned on the target audio, enabling accurate audio-driven editing. Furthermore, by feeding sequential speaker frames as context, these methods exploit the in-context learning capability of DiT-based models to generate personalized, identity-consistent talking videos.

Despite their effectiveness, mask-inpainting-based methods~\cite{zhang2024personatalk, li2024latentsync, prajwal2020lip} are highly sensitive to mask design. When assigned with small mask regions, the model tends to exploit shortcut solutions by inferring lip movements from contextual cues near the mask boundary, such as chin contours and facial dynamics, thereby bypassing audio conditioning. With large mask regions, critical background and identity-related content are masked out, leading to inaccurate background reconstruction, compromised identity consistency, and failure in challenging cases such as cuts or fast motion.

To mitigate the mask design trade-off, a feasible solution~\cite{peng2025omnisyncuniversallipsynchronization} is to adopt mask-free training on collected paired data~\cite{10.1007/978-3-030-58589-1_42} with consistent identity, pose, and background but varying lip movements. This help the model to learn both lip editing and visual consistency. However, such data requires highly controlled recording conditions and is difficult to scale, limiting practicality and generalization.

In this work, we propose Progressive Self-Correction (PSC), a mask-free lip-syncing framework comprising two stages. In the first stage, we train a multi-reference, mask-inpainting-based audio-to-lip model and distill it into a few-step diffusion model for efficient generation of paired data (with same identity and background but different lip movements). In the second stage, we use the Stage 1 model to generate synthetic pairs on-the-fly and train a mask-free lip-editing model from these pseudo-paired samples. Since the pseudo videos generated by the Stage 1 model still suffer from artifacts near facial boundaries and background distortion due to masking, we introduce a fusion module to correct the training data by replacing the generated background with that of the original video. This ensures the model learns to edit lips accurately while preserving background fidelity and identity consistency.

Our main contributions are as follows:

\begin{itemize}
    \item We identify the mask-induced trade-off in lip-syncing and propose Progressive Self-Correction, enabling a transition from mask-based to mask-free editing via self-generated data.
    \item We introduce an efficient online pipeline for generating pseudo-paired data with consistent background and different lip movements.
    \item Building on the above, we introduce SyncAnyone, a novel lip-syncing framework that robustly handles diverse in-the-wild scenarios and achieves state-of-the-art performance in terms of generation quality, temporal consistency, and motion naturalness.
\end{itemize}

\section{Related Works}
\label{sec:related}

Audio-driven lip synchronization aims to edit an existing video to align the lip movements of a subject with a target audio track, while preserving all other content, such as head pose, identity, and background. This task is framed as a video-to-video editing problem. Following the evolution of their core technical approaches, we classify existing methods into two main categories: GAN-based methods, and emerging diffusion-based methods.
\begin{figure*}[t]
    \centering
    \includegraphics[width=\textwidth]{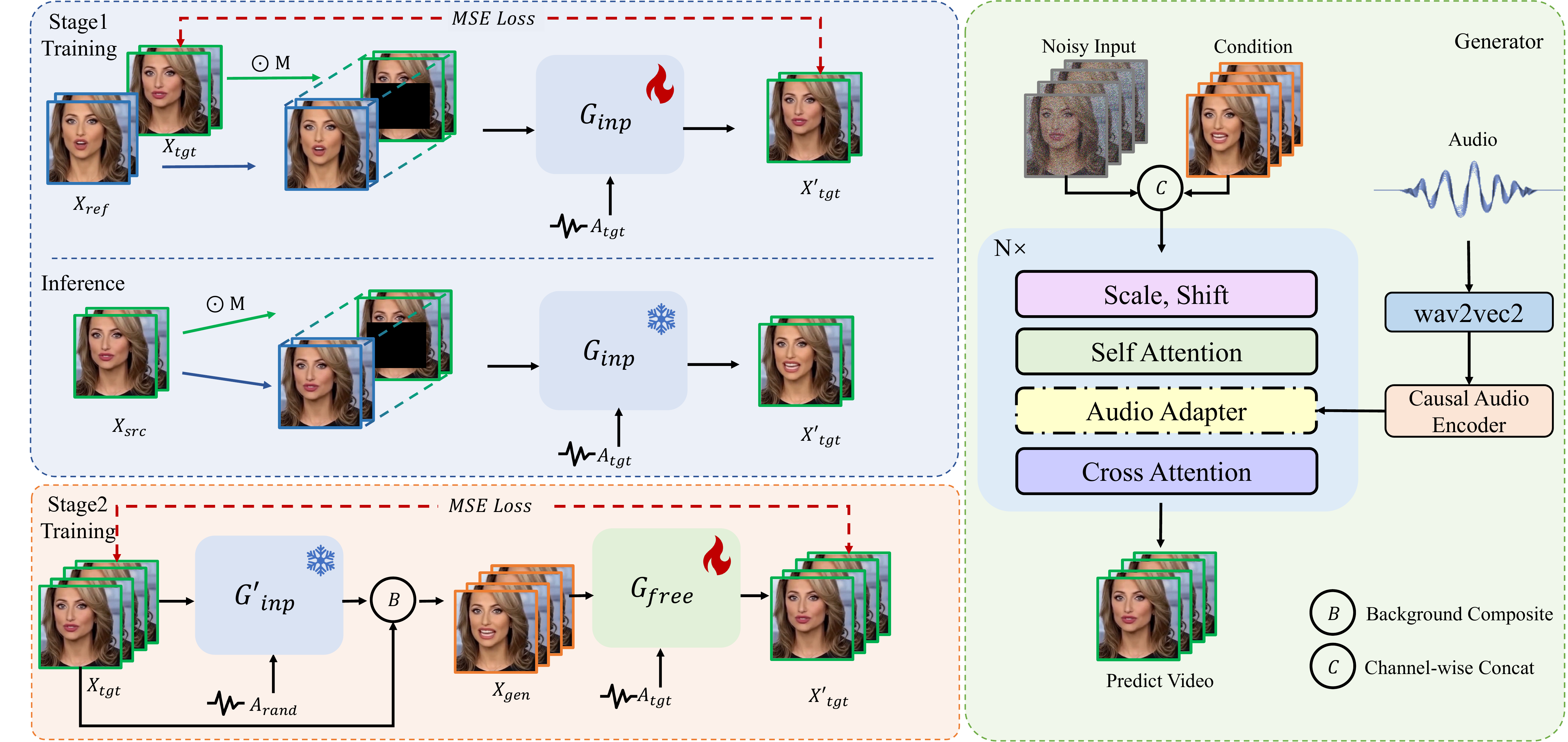}
    \caption{The overall framework of SyncAnyone. The left panel illustrates our two-stage Progressive Self-Correction (PSC) training paradigm for modifying a source video's lip movements to match a target audio. In Stage 1, a multi-reference mask inpainting model is trained for robust synthesis. In Stage 2, this model is leveraged to create a pseudo-paired dataset, which in turn supervises the training of our final, efficient mask-free model. The right panel details the specific network architecture employed in our framework.}
    \label{fig:framework}
\end{figure*}

\subsection{GAN-based Methods}
Early and mainstream lip synchronization methods predominantly rely on Generative Adversarial Networks (GANs)~\cite{goodfellow2020generative, doukas2021headgan, guo2024liveportrait}. Wav2Lip~\cite{prajwal2020lip} stands as a landmark work in this field. It pioneered the use of a pre-trained lip-sync discriminator, SyncNet~\cite{chung2017out}, to supervise the generator's training, which significantly enhanced the accuracy of the generated lip movements and established a benchmark for subsequent research.

Building on this foundation, subsequent works have introduced improvements from various perspectives. Some methods focus on enhancing generation quality and identity preservation. For instance, DINet~\cite{zhang2023dinet} deforms the feature maps of a reference image to generate more natural mouth shapes, thereby better preserving high-frequency details. VideoReTalking~\cite{cheng2022videoretalking} decomposes the task into three stages: semantic-guided reenactment, lip synchronization, and identity-aware refinement. StyleSync~\cite{guan2023stylesync} employs StyleGAN2~\cite{karras2020analyzing} as its generator backbone, leveraging its powerful generative capabilities to improve visual quality. Most of these methods are arbitrary-subject models, designed to build a general model that does not require retraining for different identities.

Another line of work introduces intermediate representations to guide the generation process. For example, some works first predict facial landmarks~\cite{xie2021towards, lu2021live} or 3D-reconstruction-based landmarks~\cite{song2022everybody} from audio before performing image-to-image translation. IP-LAP~\cite{zhong2023identity} adopts such a two-stage approach. Some methods~\cite{guan2024resyncer, zhang2024personatalk} further incorporates 3D mesh priors to guide facial motion, effectively reducing artifacts. Additionally, personalized methods~\cite{obama, NVP, adnerf} exist, which can achieve photorealistic results by training on a specific person. However, they suffer from poor generalization and limited applicability.

\subsection{Diffusion-based Methods}
More recently, the advent of Diffusion Models has led to a significant leap in performance for audio driven portrait animation~\cite{tian2024emo, meng2025mirrorme, li2025infinityhuman, gao2025wan, yang2025infinitetalk}, 
which makes it feasible to perform lip-syncing tasks using diffusion models.
with these methods typically generating results of higher resolution and richer detail. They can be broadly categorized into end-to-end and two-stage architectures.

End-to-end models directly generate target video frames conditioned on audio. LatentSync~\cite{li2024latentsync} is an end-to-end framework based on an audio-conditioned Latent Diffusion Model (LDM), which requires no intermediate motion representation. SayAnything~\cite{ma2025sayanything} and Diff2Lip~\cite{mukhopadhyay2024diff2lip} follow a similar idea, performing audio-conditioned denoising generation directly in the latent or pixel space.

Two-stage models decompose the task into audio-to-motion and motion-to-image steps. For example, MyTalk~\cite{yu2024make} uses a diffusion model in the first stage to convert audio to motion and a VAE in the second stage for image generation. StyleSync (diffusion version)~\cite{zhong2024style} employs a Transformer in its first stage and a diffusion model in its second. DiffDub~\cite{liu2024diffdub} utilizes a diffusion autoencoder to convert masked images into semantic latent codes, which are then used alongside audio to condition the final image generation.

To break through the limitations of this paradigm, recent work has begun to explore more universal solutions. For instance, OmniSync~\cite{peng2025omnisyncuniversallipsynchronization} pioneered a mask-free framework. It employs a Diffusion Transformer (DiT) to directly edit video frames, aiming to move beyond the dependency on facial priors and extend applicability to arbitrary AI-generated content (AIGC).

In summary, the field of audio-driven lip synchronization has progressed significantly, evolving from GAN-based frameworks to diffusion-based ones, achieving substantial gains in generation quality and realism. However, existing works, regardless of their underlying architecture, predominantly rely on a common paradigm: masking and inpainting of the mouth region, introduces several inherent and difficult-to-overcome limitations: limited head pose, difficulty in identity preservation, boundary artifacts.

\section{Method}
\label{sec:method}

Our goal is to create a lip-sync model that generates high-fidelity, background-consistent results with high efficiency. To achieve this, we introduce Progressive Self-Correction (PSC), a two-stage learning paradigm designed to systematically decouple motion robustness from background preservation. As illustrated in Fig.~\ref{fig:framework}, PSC first synthesizes a motion-correct but potentially flawed video (Stage 1), and then uses it to supervise a final, highly efficient mask-free model that learns to correct these flaws while maintaining pixel-perfect background integrity (Stage 2).

\subsection{Preliminaries}

Our framework is built upon a synergistic combination of state-of-the-art generative modeling techniques, which we briefly outline below.

\noindent\textbf{Flow Matching.}
We adopt Flow Matching~\cite{lipman2022flow} as our core generative paradigm. It directly learns a deterministic vector field $v_t$ that transports a prior distribution (e.g., $\mathcal{N}(0, I)$) to the data distribution. The model, parameterized by $\theta$ and conditioned on a context $c$, is trained by regressing this vector field via a simple L2 objective:
\begin{equation}
    \begin{aligned}
        \mathcal{L}_\theta = 
        \mathbb{E}_{t, x_0, x_1, c} \left[ \| v_t((1-t)x_0 + tx_1, c; \theta) - (x_1 - x_0) \|_2^2 \right],
        \label{eq:flow_matching_base}
    \end{aligned}
\end{equation}

\noindent where $x_0$ and $x_1$ are samples from the prior and data distributions, respectively. This approach offers a more stable and efficient training path compared to traditional diffusion models.

\noindent\textbf{Diffusion Transformer (DiT).}
Our network backbone is the Diffusion Transformer (DiT), which replaces the common U-Net with a Transformer operating on latent video patches (tokens). Its self-attention mechanism is highly effective at capturing long-range spatiotemporal dependencies. Conditioning information, such as the timestep $t$ and other contexts, is efficiently integrated into each block via adaptive layer normalization schemes (adaLN).

\noindent\textbf{I2V-inspired Input Structure.}
To manage our complex conditional inputs, our model's architecture is inspired by modern Image-to-Video (\textbf{I2V}) frameworks. These models are designed to generate video sequences conditioned on a source image, using a unified input structure for a Transformer backbone. This structure typically consists of three parallel channels that are concatenated before being fed to the network: the noisy latents $Z_t$, a conditioning channel $Y$, and a binary mask channel $M$.

The input to the DiT backbone at each timestep $t$ can be formally expressed as:
\begin{equation}
    Z_{\text{in}} = \text{Concat}([Z_t, M, Y], \text{dim=channel}),
    \label{eq:i2v_in}
\end{equation}

\noindent here, $Z_t$ represents the noisy latents of the target video frames to be generated. The $Y$ channel carries the conditioning information; in an I2V context, this is typically the latent representation of the source image, replicated across the temporal dimension. The $M$ channel is a binary mask that spatially or temporally guides the generation. For instance, a value of 1 in the mask can indicate a reference frame to be conditioned upon, while 0 indicates a frame to be generated. This powerful and flexible input structure allows the model to learn complex generation tasks by interpreting the relationships between these parallel input streams. Our work adapts this structure for the specific demands of the lip-syncing task.

\subsection{Stage 1: Lip Sync via Mask Inpainting }

The objective of Stage 1 is to create a mask inpainting model, $G_{inp}$, that is robust to large head poses and can generate dynamically plausible lip movements. We adapt the I2V-inspired architecture for this task, specializing its input channels for lip-syncing.

\noindent\textbf{Input Channel Specialization and Training.}

We adapt the I2V-inspired input structure for the mask-inpainting task by specializing its parallel channels. Let $X_{tgt} = \{x_1, \dots, x_k\}$ denote a sequence of ground-truth target frames sampled from the dataset, which corresponds to the target audio $A_{tgt}$. To guide the synthesis, we define a spatial mask $M \in \{0, 1\}^{H \times W}$, where a value of 1 indicates the background regions to be preserved, and 0 indicates the mouth region to be inpainted (typically a large rectangular area over the lower face). The conditioning frames $X_{cond}$ are formally derived by masking the ground-truth frames:
\begin{equation}
    X_{cond} = X_{tgt} \odot M,
\end{equation}
where $\odot$ denotes the element-wise (Hadamard) product. 

The model is trained end-to-end within the Flow Matching paradigm, taking $X_{cond}$, $M$, and $A_{tgt}$ as joint conditions. The training objective is to predict a vector field that reconstructs the original $X_{tgt}$ by synthesizing high-fidelity lip movements within the masked regions that are synchronized with $A_{tgt}$. Formally, the generation of the predicted frames $X'_{tgt}$ is expressed as:
\begin{equation}
    X'_{tgt} = G_{inp}(X_{cond}, M, A_{tgt}).
\end{equation}
During training, the network minimizes the reconstruction error between the generated $X'_{tgt}$ and the ground-truth $X_{tgt}$, forcing the model to learn the spatial context from the unmasked areas while following the phonetic cues from the audio.

\noindent \textbf{Multi-Reference Temporal Modeling.} 
To capture person-specific identity and dynamic speaking styles, Stage 1 employs a multi-reference temporal modeling strategy. Unlike methods relying on a single reference image, we provide the model with a sequence of frames to exploit long-range spatiotemporal dependencies. 

During training, we process a total of $N$ frames, where the first $k$ frames serve as the reference sequence $X_{ref} = \{x_1, \dots, x_k\}$ and the remaining $N-k$ frames constitute the target sequence $X_{tgt} = \{x_{k+1}, \dots, x_N\}$. These two sequences are sampled from non-contiguous and non-overlapping segments of the same video. The conditioning sequence $X_{cond}$ is then formally constructed by temporally concatenating the reference frames with the masked target frames:
\begin{equation}
    X_{cond} = [X_{ref} \ ; \ X_{tgt} \odot M],
\end{equation}
where $M$ masks out the mouth region only for the target frames $\{x_{k+1}, \dots, x_N\}$, while the reference frames $X_{ref}$ remain fully intact to provide dense identity priors. By feeding this joint sequence into the DiT, the global self-attention mechanism naturally propagates high-fidelity texture and habitual motion features from $X_{ref}$ to the generation regions in $X_{tgt}$. To ensure the model focuses on lip synthesis, the training loss is calculated exclusively on the target sequence:
\begin{equation}
    \mathcal{L} = \mathbb{E} \| X'_{k+1:N} - X_{k+1:N} \|^2_2.
\end{equation}

During inference, given a source video $X_{src}$ to be dubbed, we use $X_{src}$ in a dual role: it acts as $X_{ref}$ to provide identity context and simultaneously serves as the masked target sequence $X_{tgt} \odot M$ to provide the driving pose and background. This ensures the generated clip preserves the original head motion and environmental consistency. For visual clarity in Fig.~\ref{fig:framework} (Stage 1), we simplify the depiction of noisy latents and mask channels; however, the network input strictly follows the channel-wise concatenation described in Eq.~\ref{eq:i2v_in}.

\noindent\textbf{Audio Feature Injection.}
To provide a precise driving signal for lip synthesis, the audio condition $A_{tgt}$ is processed and injected into the DiT backbone with careful temporal alignment. Specifically, we first employ a pre-trained Wav2Vec 2.0 model to extract high-level, information-dense feature representations from the raw audio waveform.
Inspired by MirrorMe~\cite{meng2025mirrorme}, we use an causal audio encoder with the same temporal structure of the video vae to align these audio features frame-by-frame with the compressed video latents. The features are then deeply integrated into each layer of the DiT network via cross-attention mechanisms. This allows the model to dynamically attend to the most relevant audio cues at each generation step, ensuring that the synthesized lip movements are precisely synchronized with the nuances of the speech.

\noindent\textbf{Few-Step Generation via DMD.}
A fully trained, multi-step generative model like $G_{inp}$ is too slow for online data synthesis required in Stage 2. To address this, we distill $G_{inp}$ into a highly efficient, few-step student model, $G'_{inp}$, using Distribution Matching Distillation (DMD)~\cite{yin2024onestepdiffusiondistributionmatching}. DMD is a state-of-the-art technique that converts a pre-trained flow-based model into a generator that requires only a handful of inference steps (e.g., 1-4) while preserving high generation quality. This distilled model $G'_{inp}$ serves as the core synthesis engine for our pseudo-paired data pipeline, making the entire PSC framework computationally feasible.

\subsection{Stage 2: Mask-Free Correction and Refinement}
While the Stage 1 model ($G_{inp}$) provides robust motion priors, its reliance on explicit spatial masks and multi-frame concatenation introduces an inherent mask design dilemma: a restrictive mask limits lip expressiveness, whereas an expansive mask risks corrupting identity and background details (as shown in Fig.~\ref{fig:retalk_bg}). To resolve this, Stage 2 introduces a mask-free tuning paradigm that transitions the model from explicit spatial guidance to Implicit Disentanglement. This stage optimizes a highly efficient generator, $G_{free}$, that learns to autonomously identify and modify only the mouth-related pixels by resolving cross-modal inconsistencies.

\noindent\textbf{Implicit Disentanglement via Restoration-based Supervision.}
To train $G_{free}$ without spatial masks, we require a supervisory signal that defines what to preserve and what to edit. We generate this signal via a novel pseudo-paired data pipeline (Algorithm~\ref{alg:data_pipeline}). For a ground-truth frame $x_{gt}$ and its original audio $A_{gt}$, we first synthesize an initial mouth-altered frame $x'_{gen}$ using the distilled Stage 1 model ($G'_{inp}$) conditioned on a random audio $A_{rand}$. To ensure the background remains pixel-perfect, we employ a face parser~\cite{face-parsing} to compute a union mask $M_{face}$, compositing the generated face onto the pristine background of $x_{gt}$. This yields a pseudo-pair where $x_{gen}$ (condition) and $x_{gt}$ (target) share an identical background but possess conflicting lip shapes.

\begin{algorithm}[tb]%[h!]
    \caption{Robust Pseudo-Paired Data Generation Pipeline}
    \label{alg:data_pipeline}
    \begin{algorithmic}%[1]
        \STATE \textbf{Input:} Source video frame $x_{gt}$, random audio $a_{rand}$, distilled model $G'_{inp}$, face parser $\mathcal{F}_{parse}$.
        \STATE \textbf{Output:} A pseudo-pair $(x_{gt}, x_{gen})$.
        \vspace{0.5em}
        
        \STATE {\color{gray}\textit{// Step 1: Generate initial synthetic frame with altered mouth}}
        \STATE $x'_{gen} \leftarrow G'_{inp}(x_{gt}, a_{rand})$ 
        \vspace{0.5em}
        
        \STATE {\color{gray}\textit{// Step 2: Compute union mask to robustly cover all facial regions}}
        \STATE $M'_{gen} \leftarrow \mathcal{F}_{parse}(x'_{gen})$ \COMMENT{Mask from generated frame}
        \STATE $M_{gt} \leftarrow \mathcal{F}_{parse}(x_{gt})$ \COMMENT{Mask from ground-truth frame}
        \STATE $M_{face} \leftarrow M'_{gen} \lor M_{gt}$ \COMMENT{Compute the logical OR (union) of masks}
        \vspace{0.5em}
        
        \STATE {\color{gray}\textit{// Step 3: Composite final synthetic frame with pristine background}}
        \STATE $x_{gen} \leftarrow M_{face} \odot x'_{gen} + (1 - M_{face}) \odot x_{gt}$
        \vspace{0.5em}
        
        \STATE \textbf{return} $(x_{gt}, x_{gen})$
    \end{algorithmic}
\end{algorithm}

\begin{table*}[t]
\caption{Quantitative comparison on HDTF and VFHQ.  }
\begin{center}
\resizebox{1.0\linewidth}{!}{
\begin{tabular}{lccccccccccc}
\toprule
\multirow{2}{*}{\bf Method}  & \multicolumn{5}{c}{\bf HTDF} & \multicolumn{5}{c}{\bf VFHQ} \\
\cmidrule(lr){2-6} \cmidrule(lr){7-11}
&  \multicolumn{1}{c}{FID $\downarrow$} &  \multicolumn{1}{c}{FVD $\downarrow$} & \multicolumn{1}{c}{CSIM $\uparrow$} & \multicolumn{1}{c}{LSE-C $\uparrow$} & LMD $\downarrow$ & \multicolumn{1}{c}{FID $\downarrow$} &  \multicolumn{1}{c}{FVD $\downarrow$} & \multicolumn{1}{c}{CSIM $\uparrow$} & \multicolumn{1}{c}{LSE-C $\uparrow$} & LMD $\downarrow$ \\
\midrule
Wav2Lip~\cite{prajwal2020lip}                 & 15.34 & 562.1 & 0.841 & 7.58 & 9.82  & 19.12 & 642.5 & 0.825 & 7.42 & 11.85 \\
VideoRetalking~\cite{cheng2022videoretalking} & 12.21 & 392.4 & 0.772 & 7.05 & 8.64  & 15.38 & 435.1 & 0.751 & 6.94 & 9.47  \\
IP-LAP~\cite{zhong2023identity}                & 9.82  & 338.2 & 0.795 & 7.14 & 7.85  & 11.24 & 372.6 & 0.784 & 7.03 & 8.66  \\
MuseTalk~\cite{zhang2024musetalk}              & 9.12  & 245.8 & 0.854 & 6.75 & 8.92  & 10.05 & 274.3 & 0.843 & 6.65 & 9.82  \\
LatentSync~\cite{li2024latentsync}            & 8.78  & 228.3 & 0.842 & \textbf{8.12} & 16.85 & 9.56  & 256.4 & 0.835 & \textbf{8.04} & 17.94 \\
\midrule
\textbf{Ours} ($1^{st}$ stage)               & 8.65  & 220.5 & 0.858 & 7.85 & \textbf{7.62} & 9.35  & 248.2 & 0.848 & 7.72 & \textbf{8.45} \\
\textbf{Ours} ($2^{nd}$ stage)               & \textbf{8.42} & \textbf{205.4} & \textbf{0.865} & 7.95 & 7.70 & \textbf{9.12} & \textbf{232.5} & \textbf{0.856} & 7.84 & 8.52 \\
\bottomrule
\end{tabular}}%
\end{center}
\label{tab:results1}
\end{table*}

\noindent\textbf{Mechanism of the Training Conflict.}
The core of our implicit disentanglement lies in the restoration objective: we force the model to reconstruct the pristine $x_{gt}$ from the "noisy" condition $x_{gen}$ under the guidance of the target audio $A_{gt}$. This setup creates a powerful training conflict that compels the network to learn a sophisticated tripartite mapping:

Preserve (Identity/Background): The model must recognize that the background and non-mouth facial regions in $x_{gen}$ are consistent with $x_{gt}$ and should be preserved via a "copy-and-paste" behavior.

Discard (Inconsistent Motion): It must identify that the mouth region in $x_{gen}$ (driven by $A_{rand}$) contradicts the target audio $A_{gt}$ and should be treated as undesirable noise.

Synthesize (Audio-driven Lip-sync): It must synthesize a new, synchronized mouth shape solely by attending to $A_{gt}$, seamlessly integrating it into the preserved facial context.

Through this conflict-resolution process, the network's receptive field for modification becomes purely data-driven. The model autonomously learns to disentangle static identity from dynamic motion without any predefined spatial boundaries, effectively solving the mask dilemma.

\noindent\textbf{Robustness to Synthetic Artifacts.}
A potential concern with pseudo-paired data is the propagation of blending artifacts from the background compositing module. However, our restoration paradigm is inherently self-correcting: because the target $x_{gt}$ is always a pristine, artifact-free original frame, the model learns to map "synthetic-and-potentially-flawed" inputs to "real-and-clean" outputs. Consequently, at inference time, when provided with high-quality original video frames as $x_{gen}$, the model remains robust and produces results with lossless background fidelity.

\section{Experiment}
\label{sec:experiment}

\begin{figure*}[t]
    \centering
    \includegraphics[width=0.9\linewidth]{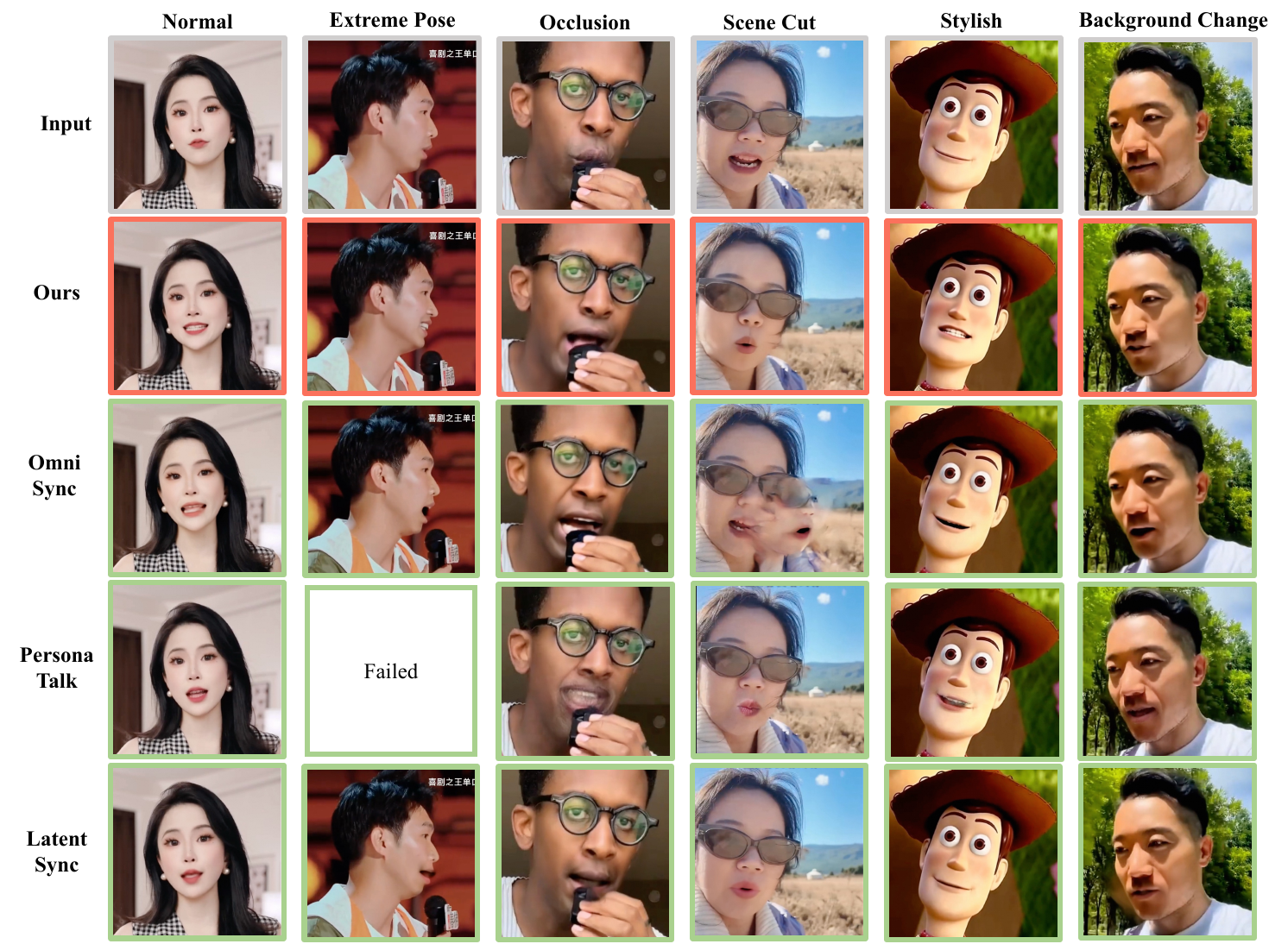}
    \caption{Qualitative comparison of our method and other methods under different scenarios. Zoom in for better visualization.}
    \label{fig:compare}
\end{figure*}

\begin{figure}[b]
    \centering
    \includegraphics[width=1.0\linewidth]{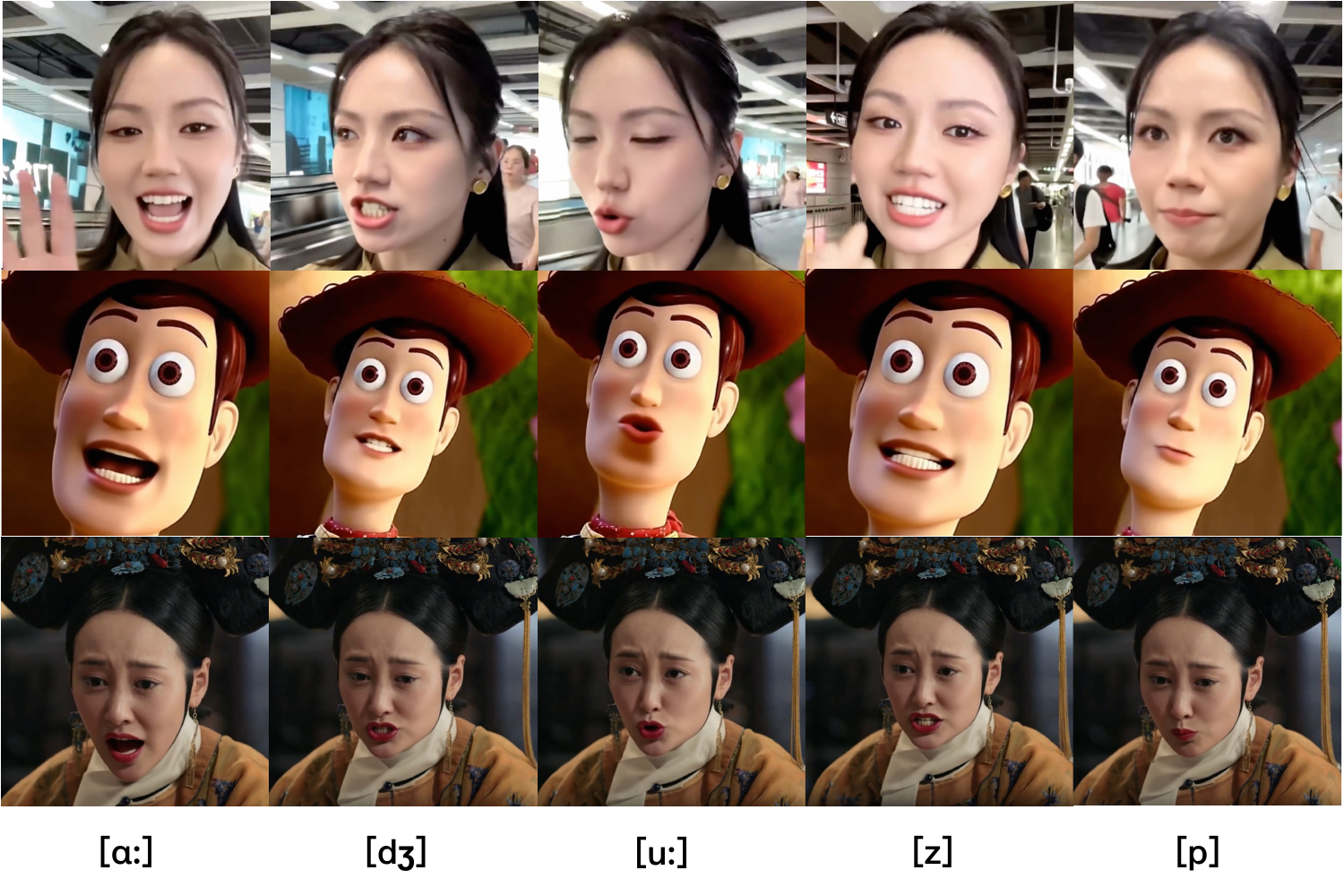}
    \caption{Lip-syncing results for different syllables.}
    \label{fig:pronounce}
\end{figure}
%-------------------------------------------------------------------------
\subsection{Experiment Settings}

\textbf{Datasets.} 

We collected approximately 100 hours of multilingual speech data, which covers scenarios such as large pose angles, occlusions, and rapid background changes.
We split these data by shots and used a face detector to extract the facial bounding boxes. Bilateral filtering was applied to these boxes to obtain smooth and compact tracking trajectories. During training, random jumps were introduced into the bounding boxes to simulate scene cuts during actual inference.

\noindent\textbf{Implementation Details}. We use the Wan2.1~\cite{wan2025} 1.3B as our base model. All models are trained on 8 NVIDIA A100 GPUs with a total batch size of 8. The training process is divided into two distinct stages:

Stage 1: In the first stage, we train the mask-inpainting model $G_{inp}$ for 100k iterations. We use the AdamW optimizer~\cite{loshchilov2019decoupledweightdecayregularization} with 
$\beta_1 = 0.9, \beta_2=0.999$, and a weight decay of $0.01$. The learning rate is initialized to $2e^{-4}$ and follows a cosine annealing schedule with a warm-up of 5k iterations.

Stage 2: The weights of the mask-free model $G_{free}$ is adopted from the first stage to keep the prior of the audio-lip syncing ablility. We train another 100k iterations with all the training parameters the same as the 1st stage. This brief yet effective fine-tuning process adapts the pre-trained generator to the new mask-free, image-to-video conditioning paradigm, resulting in a highly efficient and effective lip-syncing model.

\subsection{Experimental Comparisons}
Our evaluation is structured into two parts: a rigorous quantitative benchmark on established datasets and a qualitative analysis focused on challenging, real-world scenarios.

\noindent\textbf{Quantitative Evaluation.} To ensure a fair and comprehensive comparison, we evaluate our method against a wide range of baselines representing different technical paradigms. We conduct these experiments on two popular open-source datasets: HDTF~\cite{zhang2021flow} and VFHQ~\cite{wang2022vfhqhighqualitydatasetbenchmark}, which provide a diverse set of high-resolution talking head videos for benchmarking lip-sync accuracy and visual fidelity.

\noindent\textbf{Qualitative Evaluation.} To demonstrate the robustness of SyncAnyone in "in-the-wild" scenarios, we perform qualitative comparisons on a curated set of challenging videos featuring extreme head poses, partial occlusions, rapid background dynamics, and stylish artistic content. In these evaluations, we primarily benchmark against the top-performing SOTA methods: OmniSync, PersonaTalk, and LatentSync, focusing on their ability to maintain identity and background consistency while performing precise lip editing.

\noindent\textbf{Implementation Note.} As the source codes for OmniSync and PersonaTalk are not publicly available, we utilized their official provided web services to generate all results for comparison, ensuring that we evaluate the best possible performance of these state-of-the-art models.

\subsubsection{Quantitative Results}
We evaluate SyncAnyone on randomly sampled subsets from the HDTF~\cite{zhang2021flow} and VFHQ~\cite{wang2022vfhqhighqualitydatasetbenchmark} datasets, following established protocols to ensure a fair comparison. 
As shown in Tab.~\ref{tab:results1}, our Stage 2 model consistently achieves state-of-the-art results across visual quality, temporal stability, and identity preservation metrics. Notably, the substantial lead in FID and FVD scores highlights our model’s ability to generate high-fidelity frames with exceptional temporal coherence. Unlike mask-inpainting methods such as LatentSync~\cite{li2024latentsync} and MuseTalk~\cite{zhang2024musetalk}, which often introduce flickering or blurring artifacts near the mask boundaries, our mask-free paradigm ensures pixel-perfect background preservation and superior identity consistency, as reflected in the leading CSIM scores. In terms of synchronization, SyncAnyone maintains highly competitive LSE-C and LMD values, demonstrating that our implicit disentanglement mechanism successfully maintains precise audio-visual alignment without relying on explicit spatial constraints. Furthermore, the significant performance leap from Stage 1 to Stage 2 validates the effectiveness of our Progressive Self-Correction (PSC) framework. While Stage 1 provides robust motion priors, Stage 2 effectively eliminates mask-induced synthetic artifacts and refines high-frequency facial details. This transition results in a more realistic and stable final output, striking an ideal balance between generative realism and precise lip-synchronization.

\begin{figure}[t]
    \centering
    \includegraphics[width=1.0\linewidth]{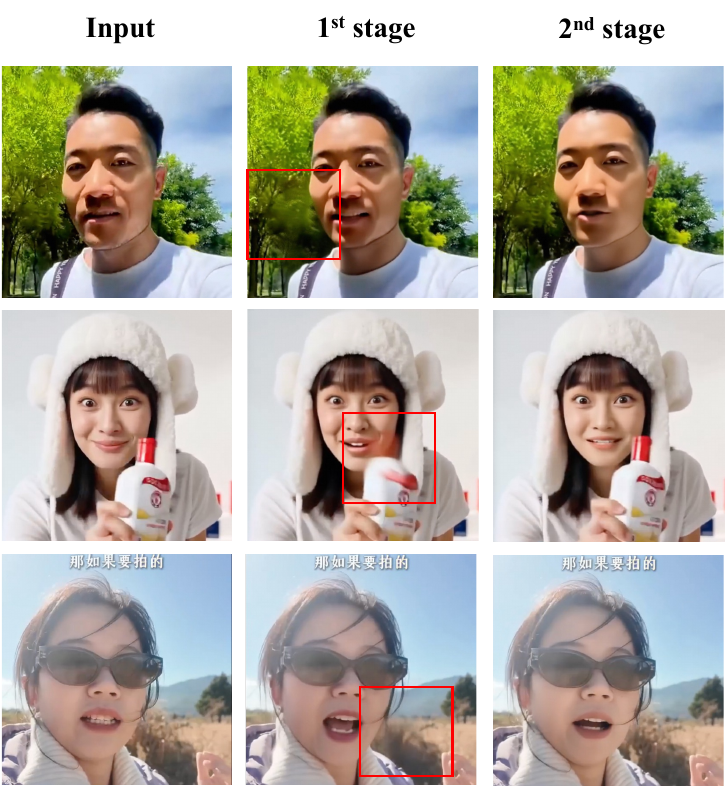}
    \caption{The comparison of the 1st and 2nd stage of our model. The regions displayed within the red box are the blurred areas in the stage-one results.}
    \label{fig:retalk_bg}
\end{figure}

\subsubsection{Qualitative Comparison}
Fig.~\ref{fig:compare} illustrates a comparative analysis between our method and state-of-the-art baselines across several challenging scenarios.

\noindent\textbf{Normal Scenario:} In frontal close-ups, our DiT-based approach produces superior lip clarity and synchronization compared to GAN and SD-based methods. As shown in Fig.~\ref{fig:pronounce}, SyncAnyone accurately captures diverse phonetic articulations with high visual fidelity.

\noindent\textbf{Extreme Pose:} At large pose angles, our method maintains profile integrity and sharp details, whereas landmark-reliant GANs often fail to recover valid 3D structures. This advantage stems from our data-driven strategy, which learns pose-specific facial dynamics directly without explicit facial priors.

\noindent\textbf{Occlusion:} Our method effectively preserves occluding objects (e.g., hands or microphones) while modifying lip movements. In contrast, GAN-based methods often generate displaced mouths due to detection failures, and inpainting-based methods like LatentSync suffer from partial occluder erosion.

\noindent\textbf{Scene Cut:} Benefiting from cut-aware data augmentation, SyncAnyone handles hard scene cuts robustly. While OmniSync often exhibits residual "ghosting" artifacts from preceding frames, our model ensures clean transitions and temporal stability.

\noindent\textbf{Stylish:} Leveraging the strong generative priors of the video transformer backbone, our model generalizes effectively to anime and artistic styles. Baselines, however, typically suffer from significant quality degradation and unnatural artifacts in these non-photorealistic domains.

\noindent\textbf{Rapid Background Change:} Thanks to our mask-free training strategy, our model achieves pixel-perfect background consistency even under high-frequency variations. Conversely, methods like OmniSync and LatentSync often produce blurred backgrounds as they fail to access or preserve complete background information during inference.

\subsubsection{Ablation Study}

To analyze the role of the two-stage training, we compared the output results of the two-stage algorithm under occlusion and background change scenarios, respectively. As shown in Fig.~\ref{fig:retalk_bg}, due to the adoption of mask inpainting, when the background or occlusion is overly complex, artifacts appear in the masked regions. In contrast, the results after the second-stage preserve the original background information while modifying the lip movements accurately, verifying the effectiveness of our method. 
The results in Tab.~\ref{tab:results1} also demonstrate that compared with the single-stage model, the two-stage model achieves superior performance on both image quality and distribution distance-related metrics, while maintaining lip motion accuracy comparable to that of mask inpainting-based methods.
\section{Conclusion}
\label{sec:conclusion}

This paper introduced SyncAnyone, a novel lip-syncing framework designed to overcome the critical limitations of mask-inpainting based editing. Our core innovation, Progressive Self-Correction, facilitates a transition from a robust mask-inpainting based model to a highly efficient mask-free one. This is achieved via a novel online pipeline that generates pseudo-paired data, teaching the model to implicitly isolate and modify the mouth region while preserving the background. As a result, SyncAnyone robustly handles challenging in-the-wild videos and achieves new state-of-the-art results in generation quality, temporal stability, and motion naturalness. We believe this self-correction paradigm offers a promising direction for future research in high-fidelity and efficient lip-syncing video synthesis.

% In the unusual situation where you want a paper to appear in the
% references without citing it in the main text, use \nocite
% \nocite{langley00}

\bibliography{example_paper}
\bibliographystyle{icml2026}

%%%%%%%%%%%%%%%%%%%%%%%%%%%%%%%%%%%%%%%%%%%%%%%%%%%%%%%%%%%%%%%%%%%%%%%%%%%%%%%
%%%%%%%%%%%%%%%%%%%%%%%%%%%%%%%%%%%%%%%%%%%%%%%%%%%%%%%%%%%%%%%%%%%%%%%%%%%%%%%
% APPENDIX
%%%%%%%%%%%%%%%%%%%%%%%%%%%%%%%%%%%%%%%%%%%%%%%%%%%%%%%%%%%%%%%%%%%%%%%%%%%%%%%
%%%%%%%%%%%%%%%%%%%%%%%%%%%%%%%%%%%%%%%%%%%%%%%%%%%%%%%%%%%%%%%%%%%%%%%%%%%%%%%
% \newpage
% \appendix
% \onecolumn
% \section{You \emph{can} have an appendix here.}

% You can have as much text here as you want. The main body must be at most $8$
% pages long. For the final version, one more page can be added. If you want, you
% can use an appendix like this one.

% The $\mathtt{\backslash onecolumn}$ command above can be kept in place if you
% prefer a one-column appendix, or can be removed if you prefer a two-column
% appendix.  Apart from this possible change, the style (font size, spacing,
% margins, page numbering, etc.) should be kept the same as the main body.
%%%%%%%%%%%%%%%%%%%%%%%%%%%%%%%%%%%%%%%%%%%%%%%%%%%%%%%%%%%%%%%%%%%%%%%%%%%%%%%
%%%%%%%%%%%%%%%%%%%%%%%%%%%%%%%%%%%%%%%%%%%%%%%%%%%%%%%%%%%%%%%%%%%%%%%%%%%%%%%

\end{document}